\documentclass[11pt,a4paper]{article}
\usepackage{acl2014}
\usepackage{times}
\usepackage{url}
\usepackage{latexsym}
\usepackage{amsmath}
\usepackage{graphicx}
\usepackage{caption}
\usepackage{booktabs}
\usepackage{mathptmx}
\usepackage{multirow}
\usepackage{subcaption}
\usepackage{framed}

\title{Two-pass Discourse Segmentation with Pairing and Global Features}

\author{Vanessa Wei Feng \\
  Department of Computer Science \\
  University of Toronto \\
  Canada \\
  {\tt weifeng@cs.toronto.edu} \\\And
  Graeme Hirst \\
  Department of Computer Science \\
    University of Toronto \\
    Canada \\
  {\tt gh@cs.toronto.edu} \\}

\date{}

\begin{document}
\maketitle
\begin{abstract}

Previous attempts at RST-style discourse segmentation typically adopt features centered on a single token to predict whether to insert a boundary before that token. In contrast, we develop a discourse segmenter utilizing a set of pairing features, which are centered on a pair of adjacent tokens in the sentence, by equally taking into account the information from both tokens. Moreover, we propose a novel set of global features, which encode characteristics of the segmentation as a whole, once we have an initial segmentation. We show that both the pairing and global features are useful on their own, and their combination achieved an $F_1$ of 92.6\% of identifying in-sentence discourse boundaries, which is a 17.8\% error-rate reduction over the state-of-the-art performance, approaching 95\% of human performance. In addition, similar improvement is observed across different classification frameworks.
\end{abstract}

\section{Introduction}
\label{sec:introduction}

Discourse parsing is the task of identifying the presence, the specific type, and possibly the hierarchical structure of discourse relations in the text. For most discourse parsers, regardless of the adopted theoretical framework, discourse segmentation, which aims to determine the boundaries of discourse units, constitutes the first stage in the pipeline workflow. Therefore, the accuracy of the discourse segmentation model is crucial to the overall performance. For parsers following Rhetorical Structure Theory (RST) \cite{Mann1988}, this first stage corresponds to segmenting the input text into non-overlapping elementary discourse units (EDUs).

Previous work on EDU segmentation has demonstrated a strong correlation between the lexico-syntactic structure and the discourse boundaries. Indeed, EDUs are frequently clauses, suggesting that EDU segmentation is just a syntactic operation.  However, \newcite{Carlson2001} have enumerated a number of exceptions to clause-based EDU segmentation. For example, clauses that are subjects or objects of a main verb are not EDUs. Therefore, the sentence 
\begin{quote}
\textit{Deciding what constitutes ``terrorism'' can be a legalistic exercise.}
\end{quote}
consists of one single EDU, instead of two EDUs segmented before \textit{can}. So simply relying on syntactic information is not sufficient for EDU segmentation, and more sophisticated approaches need to be taken. 

Recent work on RST-style discourse segmentation, including \newcite{fisher-roark:2007:ACLMain} and \newcite{xuanbach-leminh-shimazu:2012:SIGDIAL2012}, report $F_1$ scores over 90\% on identifying in-sentence boundaries. However, given the importance of the segmentation model, we seek further improvement, to help reduce error propagation to downstream components.

Previous attempts at RST-style discourse segmentation typically rely on token-centered features, i.e., features that describe the characteristics of a single token, possibly capturing some context via features such as $n$-grams, to determine whether an EDU boundary should be inserted before each particular token. In contrast, we hypothesize that both the preceding and following tokens should be equally taken into account when making decisions of whether a boundary should be inserted in between. Moreover, since individual decisions are interrelated, we hypothesize that it is helpful to incorporate features which encode global characteristics of the segmentation. We obtain these global features by performing a two-pass segmentation. Our experiments show that our pairing features as well as global features are useful for better EDU boundary recognition. Moreover, experimenting with two different segmentation frameworks, namely, sequential labeling based on Conditional Random Fields (CRFs), and a sequence independent binary classification using Logistic Regression (LR) and Support Vector Machines (SVMs), we show that the usefulness of our pairing and global features is observable across different frameworks and classifiers.

\section{Background}
\label{sec:rst}

Rhetorical Structure Theory (RST) \cite{Mann1988} is probably the most widely accepted framework for discourse study. In RST, a text is represented by a hierarchical tree structure, in which leaf nodes are EDUs, and internal nodes are larger text spans constituting multiple EDUs related by specific discourse relations, e.g., \textsc{Contrast} and \textsc{Explanation}.

\begin{figure}

[\textbf{Then, by voice vote, the Senate voted a pork-barrel bill},]${e}_1$ [\textbf{approved Thursday by the House},]${e}_2$ [\textbf{for domestic military construction}.]${e}_3$
\begin{flushright}
wsj\_0623
\end{flushright}

\centering
\includegraphics[width=0.9\columnwidth]{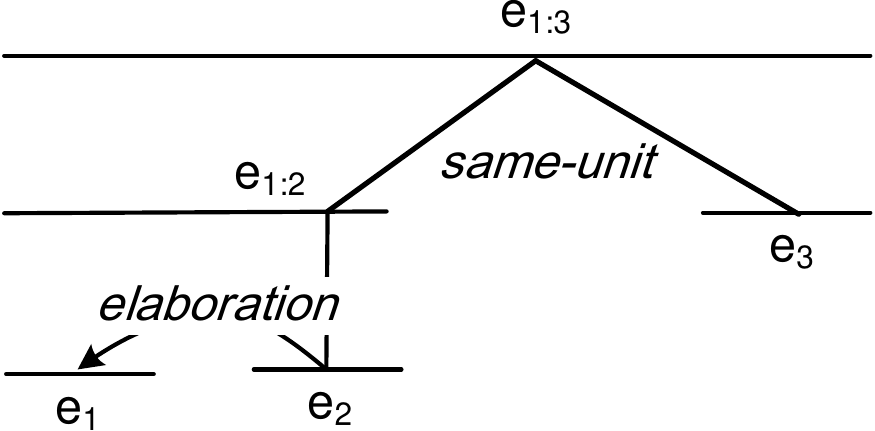}
\caption{An example sentence composed of three EDUs, with its RST discourse tree representation shown below.}
\label{fig:exampleRSTTree}
\end{figure}

For example, Figure \ref{fig:exampleRSTTree} shows a text fragment with one sentence and three EDUs. EDUs are segmented by square brackets, and the RST-style discourse tree is shown below the text. $e_1$ and $e_2$ are related by an \textsc{Elaboration} relation, where $e_1$ is more salient, called the \textit{nucleus}, while $e_2$ is called the \textit{satellite}. Then, the text span $(e_1-e_2)$ is related to $e_3$ by a \textsc{Same-Unit} relation, which is a multi-nuclear relation, in the sense that the two arguments, $(e_1-e_2)$ and $e_3$, are equally salient.

This paper is focused on the problem of EDU segmentation. The problem is formulated as finding proper EDU boundaries in the text, and extracting the token sequence in between two adjacent boundaries as one single EDU. These EDUs serve as the bottom-level discourse units in an RST-style discourse tree, which is the ultimate output of a discourse parser. Therefore, EDU segmentation is the first stage of RST-style discourse parsing. 

\section{Related Work}
\label{sec:relatedWork}

Conventionally, the task of automatic EDU segmentation is formulated as: given a sentence, the segmentation model identifies the boundaries of the composite EDUs by predicting whether a boundary should be inserted before each particular token in the sentence. In particular, previous work on discourse segmentation typically falls into two major frameworks.

The first is to consider each token in the sentence sequentially and independently. In this framework, the segmentation model scans the sentence token by token, and uses a binary classifier, such as a support vector machine or logistic regression, to predict whether it is appropriate to insert a boundary before the token being examined. Examples following this framework include \newcite{Soricut:2003}, \newcite{subba2007automatic}, \newcite{fisher-roark:2007:ACLMain}, and \newcite{Joty:2012:NDF:2390948.2391047}.

The second is to frame the task as a sequential labeling problem. In this framework, a given sentence is considered as a whole, and the model assigns a label to each token, indicating whether this token is the beginning of an EDU. Conventionally, the class label $B$ is assigned to those tokens which serve as the \textit{beginning} of an EDU, and the label $C$ is assigned to other tokens. Because EDUs cannot cross sentence boundaries, the first token in the sentence is excluded in this labeling process since it is trivially the beginning of an EDU. For example, Figure \ref{fig:labeling} illustrates this sequential labeling process. The example sentence consists of 23 tokens, separated by whitespaces, and the last 22 tokens are considered in the sequential labeling process. Each token is assigned a label, $B$ or $C$, by the labeling model. If the token is labeled as $B$, e.g., the token \textit{that} and the token \textit{to} in boldface, an EDU boundary is formed before it. Therefore, the sentence is segmented into three EDUs, indicated by the square bracket pairs. A representative work following this sequential labeling framework is \newcite{Hernault:2010:SMD:2175352.2175383}, in which the sequential labeling is implemented using Conditional Random Fields (CRFs).

\begin{figure}
\begin{framed}
[ Some analysts are concerned , however , ] [ \textbf{that} Banco Exterior may have waited too long ] [ \textbf{to} diversify from its traditional export-related activities . ] (\textbf{wsj\_0616})\\

Label sequence: C C C C C C \textbf{B} C C C C C C C \textbf{B} C C C C C C C
\end{framed}
\caption{An example of a sentence with three EDUs. The tokens are separated by whitespaces and the EDUs are segmented by square brackets. The corresponding label sequence for the tokens (excluding the first token) is shown below the sentence.}
\label{fig:labeling}
\end{figure}

An interesting exception to the above two major frameworks is \newcite{xuanbach-leminh-shimazu:2012:SIGDIAL2012}'s reranking model, which obtains the best segmentation performance reported so far: for the $B$ class, the $F_1$ score is 91.0\% and the macro-average over the $B$ and $C$ classes is 95.1\%. The idea is to train a ranking function whose input is the $N$-best output of a base segmenter and outputs a reranked ordering of these $N$ candidates. In their work, Bach et al.\@ utilized a similar CRF-based segmenter to Hernault et al.\@'s as a base segmenter. 

Because the reranking procedure is almost orthogonal to the implementation of the base segmenter, it is worthwhile to explore the enhancement of base segmenters for further performance improvement. With respect to base segmenters, which typically adopt the two major frameworks introduced previously, the best performance is reported by \newcite{fisher-roark:2007:ACLMain}, with an $F_1$ score of 90.5\% for recognizing in-sentence EDU boundaries (the $B$ class), using three individual feature sets: basic finite-state features, full finite-state features, and context-free features.

Existing segmentation models, as introduced in the beginning of this section, have certain limitations. First, the adopted feature sets are all centered on individual tokens, such as the part-of-speech of the token, or the production rule of the highest node in the syntactic tree which the particular token is the lexical head of. Although contextual information can be partially captured via features such as $n$-grams or part-of-speech $n$-grams, the representation capacity of these contextual features might be limited. In contrast, we hypothesize that, instead of utilizing features centered on individual tokens, it is beneficial to equally take into account the information from pairs of adjacent tokens, in the sense that the elementary input unit of the segmentation model is a pair of tokens, in which each token is represented by its own set of features. Moreover, existing models never re-consider their previous segmentation decisions, in the sense that the discourse boundaries are obtained by running the segmentation algorithm only once. However, since individual decisions are interrelated, by performing a second pass of segmentation incorporating features which encode global characteristics of the segmentation, we may be able to correct some incorrect segmentations of the initial run. Therefore, in this work, we propose to overcome these two limitations by our pairing features and a two-pass segmentation procedure, to be introduced in Section \ref{sec:methodology}.

\section{Methodology}
\label{sec:methodology}

\begin{figure}
\centering
\includegraphics[width=\columnwidth]{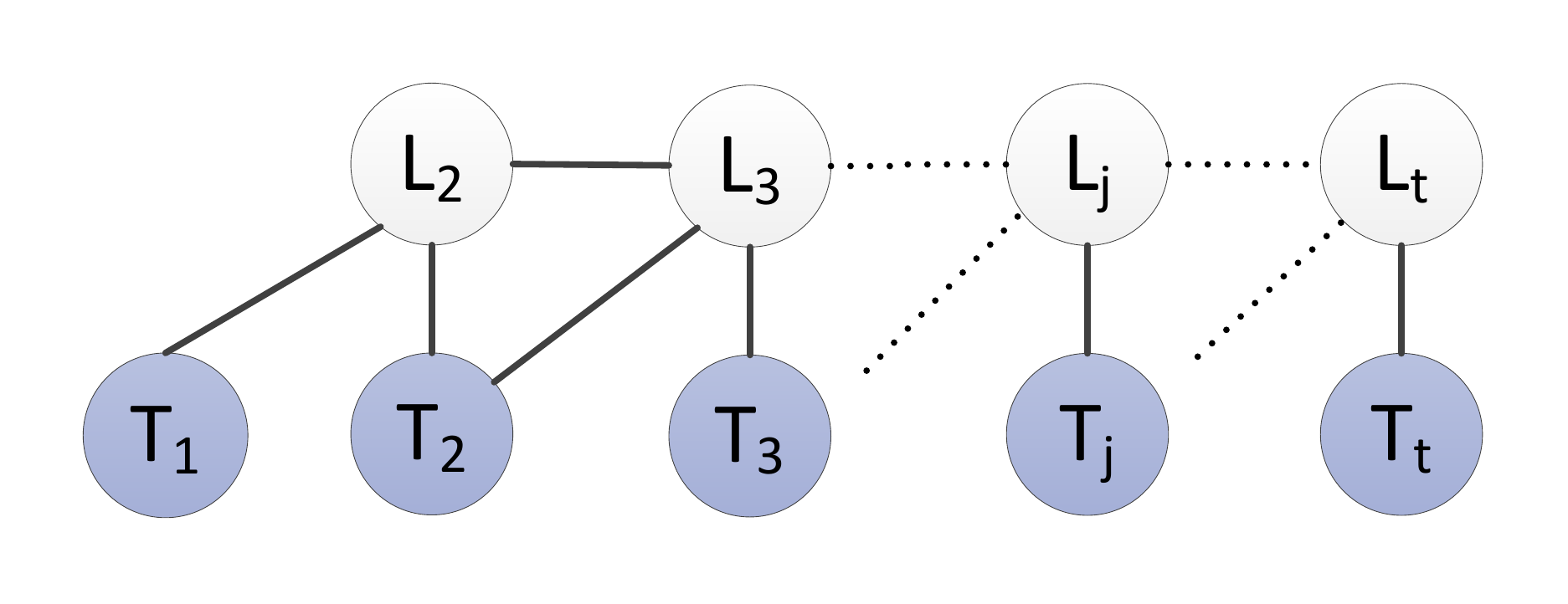}
\caption{Our segmentation model in the form of a linear-chain CRF. The first layer consists of token nodes $T_i$'s, $1 \leq i \leq t$, and the second layer represents the label $L_i$ of each pair of tokens $T_{i-1}$ and $T_i$.}
\label{fig:segmentationCRF}
\end{figure}

Figure \ref{fig:segmentationCRF} shows our segmentation model in the form of a linear-chain Conditional Random Field. Each sentence is represented by a single linear chain. For each pair of adjacent tokens in a sentence, i.e., $T_{i-1}$ and $T_i$, there is an associated binary node $L_i$ to determine the label of the pair, i.e., the existence of a boundary in between: if $L_i=B$, an EDU boundary is inserted before $T_i$; if $L_i=C$, the two adjacent tokens are considered a continuous portion in an EDU.

We choose a CRF-based model to label the whole sequence of tokens in a sentence, because a CRF is capable of taking into account the sequential information in the context, and solving the problem of determining boundaries in one single pass, which has been shown to be effective by \newcite{Hernault:2010:SMD:2175352.2175383} and \newcite{xuanbach-leminh-shimazu:2012:SIGDIAL2012}. This sequential labeling framework is also beneficial to the training process, in the sense that no additional effort needs to be made to deal with the sparsity of EDU boundaries in the data, which is usually an issue for traditional binary classifiers. 

As introduced previously, our segmentation model differs from previous work on RST-style discourse segmentation in two important ways.

First, rather than using a feature representation centered on a single token (possibly with some specifically designed features to partially incorporate contextual information), our boundary nodes take the input from a pair of adjacent tokens, to fully incorporate contextual information, allowing competition between neighboring tokens as well.

Secondly, rather than producing predictions of EDU boundaries by one pass of model application, we adopt a two-pass segmentation algorithm, which works as follows. We first apply our segmentation model once for each sentence. We then perform a second pass of segmentation, by considering some \textbf{global features} (to be described in Section \ref{sec:features}) derived from the initial segmentation. The intuition behind these novel global features is that whether a given token should be tagged as an EDU boundary sometimes depends on the neighboring EDU boundaries. For example, as suggested by \newcite{Joty:2012:NDF:2390948.2391047}, since EDUs are often multi-word expressions, the distance between the current token and the neighboring boundaries can be a useful indication. In addition, it is also helpful to know whether the tokens between the current token and the neighboring boundary form a valid syntactic constituent. Since these global indicators are available only if we have an initial guess of EDU boundaries, a second pass of segmentation is necessary. Our two-pass segmentation procedure might seem similar to Bach et al.\@'s reranking, in the sense that the segmentation takes two steps to produce final predictions. However, as discussed in Section \ref{sec:relatedWork}, in Bach et al.\@'s model, the reranking stage is almost orthogonal to the base segmentation stage, while in our two-pass model, the two stages are homogeneous and can serve as an enhancement to the base segmentation stage in the reranking framework.

%


\section{Features}
\label{sec:features}

As shown in Figure \ref{fig:segmentationCRF}, each boundary node $B_i$ in the linear-chain CRF takes the input of a pair of adjacent tokens, $T_i$ and $T_{i+1}$, in the sentence. 
Each such pair is encoded using a list of surface lexical and syntactic features, as shown in Table \ref{table:features}.

The features are partitioned into three subsets: basic features, global features, and contextual features, where the basic and contextual features are applicable for both the first and second pass, and the global features are applicable for the second pass only.

\begin{table}[t]
\centering
\begin{tabular}{p{7cm}}
\toprule
\multicolumn{1}{c}{\textbf{Basic features (for both passes)}}\\
\midrule
The part-of-speech tag and the lemma of $T_i$ / $T_{i+1}$.\\\midrule
Whether $T_i$ / $T_{i+1}$ is the beginning or the end of the sentence.\\\midrule
The top syntactic tag of the largest syntactic constituent starting from or ending with $T_i$ / $T_{i+1}$.\\\midrule
The depth of the largest syntactic constituent starting from or ending with $T_i$ / $T_{i+1}$.\\\midrule
The top production rule of the largest syntactic constituent starting from or ending with $T_i$ / $T_{i+1}$.\\
\midrule
\multicolumn{1}{c}{\textbf{Global features (for the second pass)}}\\\midrule
The part-of-speech tag and the lemma of the left / right neighboring EDU boundary.\\\midrule
The distance to the left / right neighboring EDU boundary.\\\midrule
Number of syntactic constituents formed by the tokens between $T_i$ / $T_{i+1}$ and  the left / right neighboring EDU boundary.\\\midrule
The top syntactic tag of the lowest syntactic subtree that spans from $T_i$ / $T_{i+1}$ to the left / right neighboring EDU boundary.\\
\midrule
\multicolumn{1}{c}{\textbf{Contextual features (for both passes)}}\\\midrule
The previous and the next feature vectors.\\
\bottomrule
\end{tabular}
\caption{The list of features used in segmentation.}
\label{table:features}
\end{table}

\section{Comparison with Other Models}
\label{sec:comparisonWithBaselines}

We first study how our proposed two-pass discourse segmenter based on pairing features performs against existing segmentation models. In this experiment, we train our linear-chain CRF models on the RST Discourse Treebank (RST-DT) \cite{Carlson2001}, which is a large discourse corpus annotated in accordance with RST. By convention, the corpus is partitioned into a training set of 347 documents and a test set of 38 documents. The detailed characteristics of the corpus are shown in Table \ref{table:characteristics}.

\begin{table}
\centering
\begin{tabular}{lrr}
\toprule
& \textbf{Training} & \textbf{Test}\\
\midrule
\# of documents	& 347 & 38\\
\# of sentences	&	7,455	& 992\\
\# of EDUs	&	18,765	&	2,346\\
\# of in-sentence boundaries	&	11,310	&	1,354\\
\bottomrule
\end{tabular}
\caption{Characteristics of the training and the test set in RST-DT.}
\label{table:characteristics}
\end{table}

The data are preprocessed using Charniak and Johnson's reranking parser \cite{charniak-johnson:2005:ACL} to obtain syntactic structures. Our linear-chain CRFs are designed using CRFSuite \cite{CRFsuite}, which is a fast implementation of linear-chain CRFs.

To apply our two-pass segmentation strategy (introduced in Section \ref{sec:methodology}), we first train our model by representing each sentence with a single linear chain, using the basic features and the contextual features as shown in Table \ref{table:features}. For both the training and the test set, we apply the trained one-pass model to obtain an initial EDU segmentation for each sentence. We then derive global features from this initial segmentation, and train our second-pass CRF model, together with the basic and the contextual features.

Two evaluation methods have been used in this task: the first is to evaluate the precision, recall, and $F_1$ scores of retrieving the in-sentence boundaries (the $B$ class), which is the class that we care more about. The second is to evaluate the performance of both the two classes, $B$ and $C$, based on the macro-averaged precision, recall, and $F_1$ scores of retrieving each class.

\begin{table}[t]
\centering
\begin{tabular}{lccc}
\toprule
\textbf{Model} & \textbf{Precision} & \textbf{Recall} & \textbf{$F_1$ score}\\
\midrule
CRFSeg & 91.5 & 87.6 & 89.5\\\midrule
SPADE & 85.4 & 84.1 & 84.7\\
S\&E & 85.6 & 86.6 & 86.1\\
Joty et al.\@ & 88.0 & \textbf{92.3} & 90.1\\
F\&R & 91.3 & 89.7 & 90.5\\\midrule
Reranking & 91.5 & 90.4 & 91.0\\\midrule
Ours & \textbf{92.8} & \textbf{92.3} & \textbf{92.6}\\\midrule
\textit{Human} & \textit{98.5} & \textit{98.2} & \textit{98.3}\\
\bottomrule
\end{tabular}
\caption{Performance (\%) of our two-pass segmentation model in comparison with other baseline models and human performance, evaluated on the $B$ class.}
\label{table:performanceBClass}
\end{table}

Table \ref{table:performanceBClass} demonstrates the performance evaluated on the $B$ class. We compare against several other segmentation models. In the first section, \textbf{CRFSeg} \cite{Hernault:2010:SMD:2175352.2175383} is a model that adopts a similar CRF-based sequential labeling framework as ours, but with no pairing and global features involved. The second section lists four previous works following the framework of independent binary classification for each token, including \textbf{SPADE} \cite{Soricut:2003}, \textbf{S\&E} \cite{subba2007automatic}, \textbf{Joty et al.}\@ \cite{Joty:2012:NDF:2390948.2391047}, and \textbf{F\&R} \cite{fisher-roark:2007:ACLMain}. The last model, \textbf{Reranking} \cite{xuanbach-leminh-shimazu:2012:SIGDIAL2012}, implements a discriminative reranking model by exploiting subtree features to rerank the $N$-best outputs of a base CRF segmenter, and obtained the best segmentation performance reported so far\footnote{The results of the baseline models in Tables \ref{table:performanceBClass} and \ref{table:performanceBCClasses} are the ones originally reported in the papers cited, not from our re-implementation.}. As can be seen, in comparison to all the baseline models, our two-pass model obtains the best performance on all three metrics across two classes. In fact, we obtain the same recall as Joty et al.\@, but our precision is noticeably higher than theirs. With respect to the $F_1$ score, our model achieves an error-rate reduction of 17.8\% over the best baseline, i.e., \textbf{Reranking}, and approaches to level of 95\% of human performance on this task\footnote{The human performance is measured by \newcite{Soricut:2003}.}. Moreover, since the reranking framework of Bach et al.\@ is almost orthogonal to our two-pass methodology, in the sense that our two-pass segmentation model can serve as a stronger base segmenter, further improvement can be expected by plugging our two-pass model into the reranking framework.

\begin{table}
\centering
\begin{tabular}{lcccc}
\toprule
\textbf{Model} & \textbf{Class} & \textbf{Prec} & \textbf{Rec} & \textbf{$F_1$}\\
\midrule
\multirow{3}{*}{CRFSeg} & $B$ & 91.5 & 87.6 & 89.5\\
 & $C$ & 99.0 & 99.3 & 99.2\\
 \cmidrule{2-5}
 & Macro-Avg & 95.2 & 93.5 & 94.3\\
 \midrule
\multirow{3}{*}{Reranking} & $B$ & 91.5 & 90.4 & 91.0\\
 & $C$ & 99.3 & 99.4 & 99.2\\
 \cmidrule{2-5}
 & Macro-Avg & 95.4 & 94.9 & 95.1\\
 \midrule
\multirow{3}{*}{Ours} & $B$ & \textbf{92.8} & \textbf{92.3} & \textbf{92.6}\\
 & $C$ & \textbf{99.5} & \textbf{99.5} & \textbf{99.5}\\
 \cmidrule{2-5}
 & Macro-Avg & \textbf{96.1} & \textbf{95.9} & \textbf{96.0}\\
\bottomrule
\end{tabular}
\caption{Performance (\%) of our two-pass segmentation model in comparison with other segmentation models and human performance, evaluated on the $B$ and $C$ classes and their macro-average.}
\label{table:performanceBCClasses}
\end{table}

Table \ref{table:performanceBCClasses} demonstrates the performance evaluated on both classes and their macro-average. Only two baseline models, CRFSeg and Reranking, reported their performance based on this evaluation, so other models are not included in this comparison. As can be seen, among the three models considered here, our two-pass segmentation model with pairing features performs the best not only on the $B$ class but also on the $C$ class, resulting in a macro-averaged $F_1$ score of 96.0\%.

\section{Error Propagation to Discourse Parsing}
\label{sec:inParsing}

As introduced in Section \ref{sec:introduction}, discourse segmentation is the very first stage in an RST-style discourse parser. Therefore, it is helpful to evaluate how the overall performance of discourse parsing is influenced by the results of different segmentation models.

\begin{table}
\begin{tabular}{llccc}
\toprule
Level & Segmentation & Span & Nuc & Rel\\\midrule
 \multirow{3}{*}{Intra} & Joty et al.\@ & 78.7 & 70.8 & 60.8\\
 & Ours & \textbf{85.1} & \textbf{77.5} & \textbf{66.8}\\
 \cmidrule{2-5}
 & Manual & 96.3 & 87.4 & 75.1\\\midrule
 \multirow{3}{*}{Multi} & Joty et al.\@ & \textbf{71.1} & 49.0 & 33.2\\
 & Ours & \textbf{71.1} & \textbf{49.6} & \textbf{33.7}\\ 
 \cmidrule{2-5}
 & Manual & 72.6 & 50.3 & 34.7\\\midrule
 \multirow{3}{*}{Text} & Joty et al.\@ & 75.4 & 61.7 & 49.1\\
 & Ours & \textbf{78.7} & \textbf{64.8} & \textbf{51.8}\\
 \cmidrule{2-5}
 & Manual & 85.7 & 71.0 & 58.2\\
\bottomrule
\end{tabular}
\caption{The result of discourse parsing using different segmentation. The performance is evaluated on intra-sentential, multi-sentential, and text level separately, using the unlabeled and labeled F-score.}
\label{table:inParsing}
\end{table}

We use the state-of-the-art RST-style discourse parser \cite{feng-hirst:2014:ACL2014} as the target discourse parser, and feed the parser with three sets of EDUs: (1) \textbf{manual}: the gold-standard EDUs as in the annotation of RST-DT; (2) \textbf{Joty et al.}, which is the EDUs segmented using the released version of \newcite{Joty:2012:NDF:2390948.2391047}'s segmenter\footnote{The code is vailable at \url{http://alt.qcri.org/discourse/Discourse_Parser_Dist.tar.gz}. Note that, in this released version, sentence splitting is incorporated as part of the preprocessing procedure of the software. For the sake of fair comparison, to rule out the complication of different sentence splitting between their software and our own models, we modified their code to ensure all EDU segmenters are fed with the same set of sentences as input.}; and (3) \textbf{ours}: the EDUs segmented using our own model. 

To evaluate the performance, we use the standard unlabeled and labeled F-score for Span, Nuclearity, and Relation, as defined by \newcite{MarcuBook2000}. Moreover, to further illustrate the effect of automatic segmentation on different levels of the text, we conduct the evaluation on the intra-sentential, multi-sentential, and text level separately. On the intra-sentential level, the evaluation units are discourse subtrees which do not cross sentence boundaries. On the multi-sentential level, all discourse subtrees which span at least two sentences are considered. On the text level, all discourse subtrees are evaluated.

The results are shown in Table \ref{table:inParsing}. As can be seen, on the intra-sentential level, the influence of segmentation is significant. Evaluated on Span, Nuclearity, and Relation, using our own segmentation results in a $~10$\% difference in F-score ($p <.01$ in all cases)\footnote{All significance tests are performed using the Wilcoxon signed-rank test.}, while the difference is even larger when using Joty et al.\@'s segmentation. Nonetheless, the overall parsing performance is significantly better ($p<.01$) when using our segmentation model than using Joty et al.\@'s.

However, the difference between using manual and automatic segmentation almost disappears when evaluated on multi-sentential level. In fact, the absolute difference on all metrics is less than 1\% and insignificant as well. Actually, this is not a surprising finding: Most discourse constituents in an RST-style discourse parse tree conform to sentence boundaries. Moreover, the target discourse parser we adopt in this experiment takes a two-stage parsing strategy: in the first stage, sentences are processed to form sentence-level discourse subtrees, which in turn serve as the basic processing unit in the second parsing stage. Therefore, due to the nature of the RST-style discourse trees and the particular parsing algorithm in the target discourse parser, the influence of different segmentations is very much confined within sentence boundaries, and thus has little effect on higher levels of the tree. Based on the analysis, the influence of segmentation on the text level is almost entirely attributed to its influence on the intra-sentential level.

\section{Feature Analysis}
\label{sec:featureFrameworkAnalysis}
In this section, we study the effect of our pairing and global features, the two distinct characteristics of our two-pass segmentation model, on the overall performance, and their generality across different segmentation frameworks.

\subsection{Feature Ablation across Different Frameworks}
First, starting from our full model, we perform a series of feature ablation experiments. In each of these experiments, we remove one of the component features or their combinations from the feature set in training, and evaluate the performance of the resulting model. 

\begin{figure}
\centering
\includegraphics[width=0.75\columnwidth]{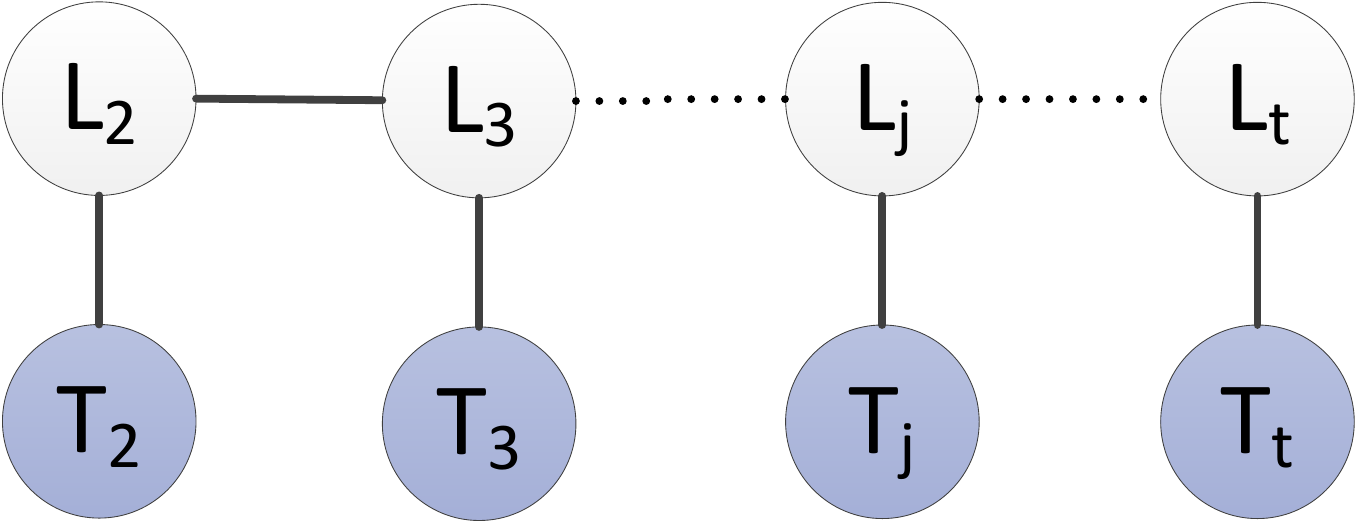}
\caption{Our segmentation model with no pairing features. The first layer consists of token nodes $T_i$'s, and the second layer represents the label $L_i$ of the single token $T_i$, representing whether an EDU boundary exists before the token.}
\label{fig:segmentationCRFNoPairing}
\end{figure}

\paragraph{Removing pairing features ($-p$)} By removing pairing features, our CRF-based segmentation model shown in Figure \ref{fig:segmentationCRF} reduces to the one shown in Figure \ref{fig:segmentationCRFNoPairing}, in which the input to each label node $L_i$ is a single token $T_i$, rather than a pair of adjacent tokens $T_{i-1}$ and $T_i$. Note that the first token $T_1$ is excluded in the sequence because there always exists an EDU boundary (the beginning of the sentence) before $T_1$. In accordance, the features listed in Table \ref{table:features} now reduce to features describing each single token $T_i$.

\paragraph{Removing global features ($-g$)} By removing global features, our two-pass segmentation model reduces to a simple one-pass model, in which only the basic and contextual features in Table \ref{table:features} are used in training the model.

\paragraph{Removing both features ($-pg$)} In this case, our model reduces to a simple one-pass model, in which only the basic and contextual features are used, and all features are based on each individual token $T_i$, rather than the token pair $T_i$ and $T_{i+1}$.

Moreover, we wish to explore the generality of our pairing features and the two-pass strategy, by evaluating their effects across different segmentation frameworks. In particular, since our two-pass segmentation model itself is a CRF-based sequential labeling model, in this experiment, we also study the effect of removing pairing and global features in the framework of independent binary classification. Recall that in the framework of independent binary classification, each token (excluding $T_1$) in a sentence is examined independently in a sequence, and a binary classifier is used to predict the label for that token. 

\begin{figure}
\centering
\begin{subfigure}[b]{0.45\columnwidth}
\centering
\includegraphics[scale=0.4]{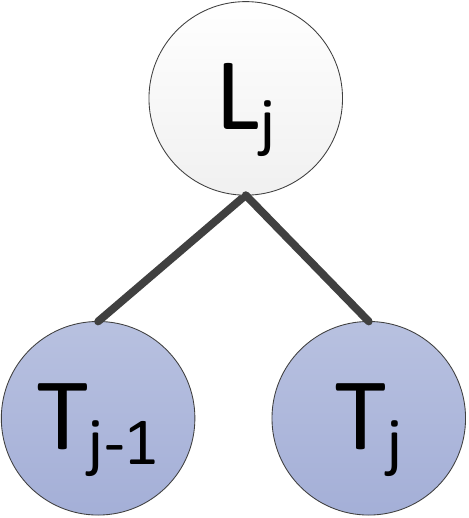}
\caption{With pairing features.}
\label{fig:binaryPairing}
\end{subfigure}
~
\begin{subfigure}[b]{0.45\columnwidth}
\centering
\includegraphics[scale=0.4]{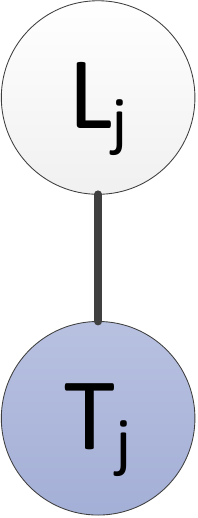}
\caption{No pairing features.}
\label{fig:binaryNoPairing}
\end{subfigure}  
\caption{Our segmentation model in the framework of independent binary classification.}
\label{fig:binaryFramework}
\end{figure}

Figure \ref{fig:binaryFramework} shows our models in the framework of independent binary classification. If pairing features are enabled, as shown in Figure \ref{fig:binaryPairing}, in each classification, a pair of adjacent tokens, rather than a single token, are examined, and the classifier predicts whether an EDU boundary exists in between. If pairing features are disabled, the model reduces to the one shown in Figure \ref{fig:binaryNoPairing}.

In this experiment, we explore two underlying classifiers in independent binary classification: Logistic Regression (LR) and a linear-kernel Support Vector Machine (SVM). We implement these two classifiers using Scikit-learn \cite{scikit-learn}. For LR, all parameters are kept to their default values, while for SVM, we use auto class-weights, which are adjusted based on the distribution of different classes in the training data, to overcome the sparsity of class $B$ in the dataset.

\begin{table}
\centering
\begin{tabular}{lccc}
\toprule
\textbf{Model} & \textbf{Precision} & \textbf{Recall} & \textbf{$F_1$ score}\\
\midrule
CRF	&	92.8 & 92.3 & 92.6\\
LR	&	\textbf{92.9} & 92.2 & 92.5\\
SVM	& 	92.6 & \textbf{92.8} & \textbf{92.7}\\
\midrule
CRF${}^{-p}$ & 91.3 & 91.1 & 91.2\\
LR${}^{-p}$ & 91.0 & 90.5 & 90.7\\
SVM${}^{-p}$ & 90.4 & 92.5 & 91.4\\
\midrule
CRF${}^{-g}$ & 92.5 & 91.0 & 91.7\\
LR${}^{-g}$ & 91.7 & 91.0 & 91.3\\
SVM${}^{-g}$ & 84.7 & 94.7 & 89.4\\
\midrule
CRF${}^{-pg}$ & 87.0 & 82.5 & 84.7\\
LR${}^{-pg}$ & 86.9 & 83.0 & 84.9\\
SVM${}^{-pg}$ & 70.5 & 94.6 & 80.8\\
\bottomrule
\end{tabular}
\caption{The effect of removing pairing features ($-p$), removing global features ($-g$), and removing both ($-pg$), in comparison with the full model in the first section, across different segmentation frameworks. CRF stands for our standard two-pass segmentation models based on linear-chain CRFs, while LR and SVM stand for two different classifiers in the framework of independent binary classification.}
\label{table:featureFrameworkAnalysis}
\end{table}

Table \ref{table:featureFrameworkAnalysis} demonstrates the results of our feature analysis. The first section lists the performance of our full model in different segmentation frameworks. As can be seen, our full models perform similarly across different frameworks, where the absolute difference in $F_1$ is less than 0.2\% and insignificant. This is consistent with \newcite{Hernault:2010:SMD:2175352.2175383}'s finding that, when a large number of contextual features are incorporated, binary classifiers such as SVM can achieve competitive performance with CRFs. The second section lists the performance of our models with no pairing features ($-p$). For all three resulting models, CRF${}^{-p}$, LR${}^{-p}$, and SVM${}^{-p}$, their performance is significantly poorer ($p < .01$) than their corresponding full model in the first section. A similar trend is observed when global features are removed ($-g$) in the third section. However, with respect to the underlying frameworks themselves, SVM is significantly worse than CRF and LR ($p < .01$), while such a significant difference is not observable when pairing features are removed. Finally, when both sets of features are removed ($-pg$), as shown in the last section, the performance of our models drops drastically (from above 90\% to below 85\%). This suggests that the pairing and global features, which have an important effect on the performance by themselves, are even more important in their combination.

In this experiment, we demonstrate that the pairing features and the global features have individual effect in improving the overall segmentation performance, and such an improvement is significant. Moreover, we observe similar effects across different frameworks, which suggests the generality of these two novel aspects of our segmentation model.

\subsection{Error Analysis}
\label{subsec:errorAnalysis}

\begin{table}
\centering
\begin{tabular}{lrrrr}
\toprule
& & \multicolumn{2}{c}{CRF}\\
\cmidrule{3-4}
& & $\neg$\textbf{Error} & \textbf{Error} & \textbf{Total}\\
\cmidrule{3-5}
\multirow{2}{*}{CRF${}^{-p}$} & $\neg$\textbf{Error} & 20357 & 52 & 20409\\
& \textbf{Error} & 100 & 149 & 249\\
\cmidrule{2-5}
& \textbf{Total} & 20457 & 201 & 20658\\
\midrule\midrule
& & \multicolumn{2}{c}{CRF}\\
\cmidrule{3-4}
& & $\neg$\textbf{Error} & \textbf{Error} & \textbf{Total}\\
\cmidrule{3-5}
\multirow{2}{*}{CRF${}^{-g}$} & $\neg$\textbf{Error} & 20436 & 0 & 20436\\
& \textbf{Error} & 21 & 201 & 222\\
\cmidrule{2-5}
& \textbf{Total} & 20457 & 201 & 20658\\
\midrule\midrule
& & \multicolumn{2}{c}{CRF${}^{-p}$}\\
\cmidrule{3-4}
& & $\neg$\textbf{Error} & \textbf{Error} & \textbf{Total}\\
\cmidrule{3-5}
\multirow{2}{*}{CRF${}^{-g}$} & $\neg$\textbf{Error} & 20339 & 97 & 20436\\
& \textbf{Error} & 70 & 152 & 222\\
\cmidrule{2-5}
& \textbf{Total} & 20409 & 249 & 20658\\
\bottomrule
\end{tabular}
\caption{Comparisons of error between our CRF-based segmentation models with different feature settings.}
\label{table:errorAnalysis}
\end{table}

\begin{figure*}
\begin{framed}
[ `` Oh , I bet ] [ it 'll be up 50 points on Monday , '' ] [ said Lucy Crump , a 78-year-old retired housewife in Lexington , Ky. ] (\textbf{CRF})\\

[ `` Oh , ] [ I bet ] [ it 'll be up 50 points on Monday , '' ] [ said Lucy Crump , a 78-year-old retired housewife in Lexington , Ky. ] (\textbf{CRF${}^{-p}$})

\end{framed}

\begin{framed}
[ They argue ] [ that the rights of RICO defendants and third parties ] [ not named in RICO indictments ] [ have been unfairly damaged . ] (\textbf{CRF})\\

[ They argue ] [ that the rights of RICO defendants and third parties ] [ not named in RICO \mbox{indictments} have been unfairly damaged . ] (\textbf{CRF${}^{-g}$})
\end{framed}
\caption{Example sentences where the full model (\textbf{CRF}) is correct while the weaker model, (\textbf{CRF${}^{-p}$}) or (\textbf{CRF${}^{-g}$}), makes mistakes.}
\label{fig:exampleErrors}
\end{figure*}

We now conduct a token-by-token error analysis to study the distributions of the errors made by our CRF-based models with different feature settings. In particular, we evaluate the labeling errors made on each token in the test set by our fully-fledged two-pass segmentation model or the models trained with pairing or global features removed. Here, we restrict our comparisons to models following the sequential labeling framework, i.e., the CRF-based models with $-p$ or $-g$ superscript in Table \ref{table:featureFrameworkAnalysis}. Once again, all tokens which are the beginning of the sentence are not included in this analysis.

The results are shown in Table \ref{table:errorAnalysis}. One interesting observation is that, as demonstrated in the second section of the table, on top of CRF${}^{-g}$, by adding global features, our full model is able to correct the 21 errors made by CRF${}^{-g}$ while introducing no additional errors in the process. In addition, as illustrated in the third section of the table, pairing and global features are almost complementary to each other, in the sense that the 39\% of the errors made by CRF${}^{-p}$ occur on cases where  CRF${}^{-g}$ is correct, and reciprocally, 32\% of errors made by CRF${}^{-g}$ happen on cases where CRF${}^{-p}$ is correct.

Finally, in Figure \ref{fig:exampleErrors}, we show some example sentences, which our fully-fledged two-pass segmentation model labels correctly, while the weaker models make some errors.


\section{Conclusion and Future Work}
\label{sec:conclusions}

In this paper, we developed a two-pass RST-style discourse segmentation model based on linear-chain CRFs. In contrast to the typical approach to EDU segmentation, which relies on token-centered features in modeling, the features in our segmenter are centered on pairs of tokens, to equally take into account the information from the previous and the following token surrounding a potential EDU boundary position. Moreover, we propose a novel two-pass segmentation strategy. After the initial pass of segmentation, we obtain a set of global features to characterize the segmentation result in a whole, which are considered in the second pass for better segmentation.

Comparing with several existing discourse segmentation models, we achieved the best performance on identifying both the boundaries and non-boundaries. Moreover, we studied the effect of our novel pairing and global features, and demonstrated that these two sets of features are both important to the overall performance, and such importance is observable across different segmentation frameworks and classifiers. Finally, we experimented with our segmentation model as a plug-in in the state-of-the-art RST-style text-level discourse parser \cite{feng-hirst:2014:ACL2014} and evaluated its influence to the overall parsing accuracy. We found that the automatic segmentation has a huge influence on the parsing accuracy, when evaluated on intra-sentential level; however, such an influence is very minor on multi-sentential level.

For future work, we wish to explore the incorporation of our two-pass segmentation model into the reranking framework of \newcite{xuanbach-leminh-shimazu:2012:SIGDIAL2012}. Since our model is shown to be a stronger base segmenter, with the reranking procedure, further improvement in segmentation accuracy may be expected.

\section*{Acknowledgments}
This work was financially supported by the Natural Sciences and Engineering Research Council of Canada and by the University of Toronto.

\bibliographystyle{acl}
\bibliography{coherencediscoursebib}

\end{document}